\documentclass[reqno]{amsart}
\usepackage{bbm}
\usepackage{amsfonts}
\usepackage{mathrsfs}
\usepackage{hyperref}
\usepackage{amssymb}
\usepackage{multirow}
\usepackage{graphicx}

\usepackage{algorithm}
\usepackage{CJK}
\usepackage{algcompatible}
 \newtheorem{thm}{Theorem}[section]

 \theoremstyle{remark}
 
 \numberwithin{equation}{section}

\begin{document}
\title[Solving a Mathematical Problem in Square War: a Go-like Board Game]
{Solving a Mathematical Problem in Square War: a Go-like Board Game}

\author{Chu Luo}
\address{Department of Computer Science and
Engineering, University of Oulu, Oulu, Finland} \email{chu.luo@ee.oulu.fi}

\thanks{We thank people who gave comments to this work.}

\begin{abstract}

In this paper, we present a board game: Square War. 
The game definition of Square War is similar to the classic Chinese board game Go. 
Then we propose a mathematical problem of the game Square War.
Finally, we show that the problem can be solved by using a method of mixed mathematics and computer science.


\end{abstract}

\maketitle \numberwithin{equation}{section}
\newtheorem{theorem}{Theorem}[section]
\newtheorem{lemma}[theorem]{Lemma}
\newtheorem{example}[theorem]{Example}
\allowdisplaybreaks

\section{Introduction}
Go, also known as Weiqi, is an old Chinese board game involving two sides, black and white. 
According the rule of Go, the board is a $19\times 19$ two-dimensional grid of straight lines. 
The game starts with a vacant board which has 361 intersections.
By convention, the black side first places a black stone, which is a game piece, on a vacant intersection.
Then two sides alternately place stones with the side color on the remaining intersections.
The complexity of Go is considered to be approximately $10^{171}$ legal states on the board, significantly more than the estimated 
number $10^{50}$ of chess \cite{tromp2007combinatorics}. 
Computational intelligence of Go still challenges computer scientists.
Unlike chess, there currently is no computer program that can defeat professional human players.

Therefore, researchers attempt to develop invincible methods in computer Go, from various perspectives
\cite{georgeot2012game}, \cite{gelly2011monte}, \cite{chaslot2006monte}, \cite{gelly2006modification},
\cite{silver2012temporal}, \cite{silver2007reinforcement} and \cite{stern2006bayesian}.
The ability of computer Go is being improved gradually as the time advances.

Furthermore, there exist variants of the game Go, where artificial intelligence programs are also developed. 
For example, the game NoGo and Go-Moku are defined using similar game settings of Go.
In \cite{alus1996go}, a computer program of Go-Moku shows that the player who moves first can always win the game through a specific strategy.
The proof is similar to the mixed proof of mathematics and computer science of the famous Four-color theorem: contiguous regions in a map can be colored with at most four colors and without any two neighbors having the same color.
A formal description of the proof of the Four-color theorem is detailed in \cite{gonthier2008formal}.
For the game NoGo, a number of studies such as \cite{chou2011revisiting} and \cite{sun2013research} introduce various methods to improve
 the ability of computer playing programs.

In this paper, we first present a board game: Square War. The game setting of Square War is very similar to Go. 
Then we propose a mathematical problem of the game Square War.
Lastly, we use a method of mixed mathematics and computer science to solve the problem.

\section{Definition of Square War}
The game Square War comprises the following rules:
\begin{enumerate} 
\item The game is played on the same chess board as Go: $19\times 19$ two-dimensional grid.
\item The game is played by two sides. 
One side uses black stones (game pieces) and the other side uses white stones.
This is also the same as Go.
\item The black side first places a stone on an empty intersection and then two sides alternately place stones on empty intersections with their side colors. However, once a stone is placed, it cannot be removed.
\item To add difficulty to the black side, the first two stones of the black side must be adjacent on a line of the grid.
\item When four stones of a side become the four vertices of a square and the edges of this square overlap the grid lines,
this side wins the game.
\end{enumerate}

\begin{figure}
\begin{center}
\includegraphics[width=0.9\textwidth]{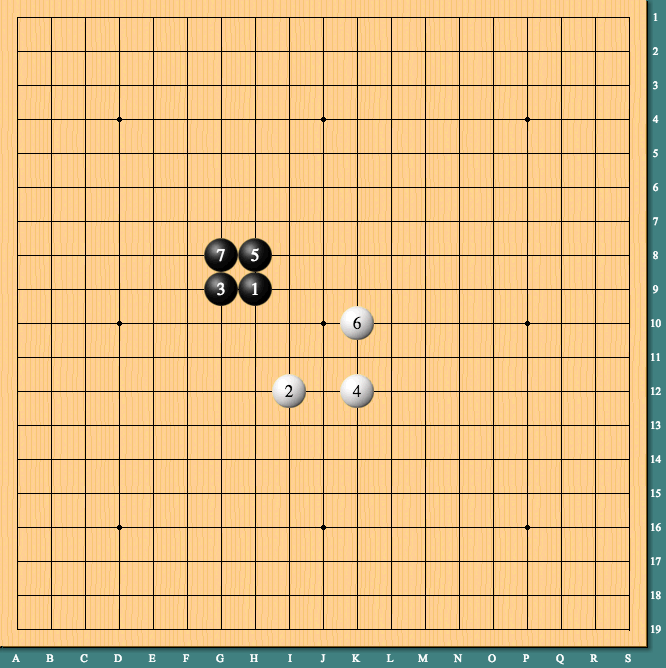}
  \caption{An example of the rules.}
  \label{f1}
  \end{center}
\end{figure}

Figure \ref{f1} gives an example of the rules where the numbers on stones represent the step sequences.
Black stones 1, 3, 5, and 7 become the four vertices of a square and the edges of this square cover the grid lines.
As a result, the black side wins. 

\section{Mathematical Problem}
Similar to other board games, the side moving first has advantage in Square War.
This advantage can be crucial. 
For example, it is mathematically proven that the player who moves first can always win the game with a tactic in Go-Moku.

In this context, we propose a mathematical problem:
\begin{itemize}
\item Can the player who moves first always win the game with a tactic regardless of the opponent tactic?
\end{itemize}

\section{Solution}
In this section, we show the answer of the problem is positive. This is equal to the following theorem.
 \begin{thm}\label{lem1.1}In the game Square War, let $X$ be the tactic set of the black side.
 Similarly, let $Y$ be the tactic set of the white side. If $x \in X$ and $y \in Y$, denote 
\begin{equation}
F(x,y) = \left\{ \begin{array}{lr} 1 & : \text{when the black side wins}\\
0 & : \text{when the white side wins.} \end{array} \right.
\end{equation}
$\forall y \in Y$, there always exists $x \in X$ such that $F(x,y) = 1$.
  \end{thm}

\begin{figure}
\begin{center}
\includegraphics[width=0.9\textwidth]{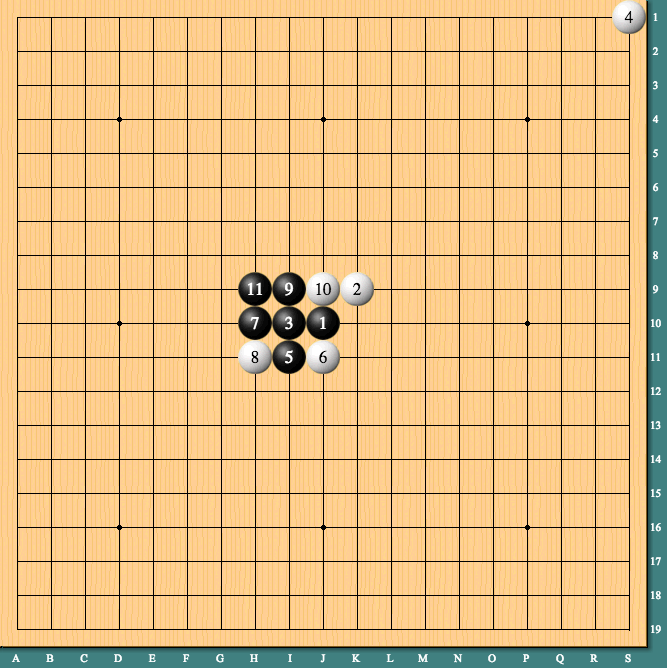}
  \caption{A possible situation of Square War}
  \label{f2}
  \end{center}
\end{figure}

\begin{proof}
We prove with mixed mathematics and computer concepts.
Given $y$, it is sure that the brute-force search can verify whether there is a related $x$, as the legal states on the board is finite.
However, the computational complexity of the search may be extremely high.
We use an algorithm $T$ to give a $X_0$ where $X_0 \subset X$. 
And we show that $\forall y \in Y$, there always exists $x \in X_0$ such that $F(x,y) = 1$.

In $T$, $\forall x \in X_0$ starts with the position of $(J, 10)$ on the board, which is the centre of the board.
After the centre is captured by the black side, we can assume that $\forall y \in Y$ starts with the position
of $(p,q)$ where $p$ is from $K$ to $S$ and $1 \leq q \leq 10$, 
because the board is symmetric about the stone $1$, the line $J$ and the line $10$.
Then, $T$ assigns the next black stone of $\forall x \in X_0$ to $(I,10)$.

The fourth step is critical now. 
We define a set $U$ by stating that $\forall u \in U$, the triple of stone $2$, $(J,11)$ and $u$ can form an isosceles right triangle which has two edges overlap the grid lines of the board.
Similarly, we define a set $V$ by stating that $\forall v \in V$, the triple of stone $2$, $(H,11)$ and $v$ can form an isosceles right triangle which has two edges overlap the grid lines of the board.
We then define a set $W=\{(I,11)\}\cup\{(J,11)\}\cup\{(H,10)\}\cup\{(H,11)\}\cup\{(I,9)\}\cup\{(J,9)\}\cup\{(H,9)\}\cup\{(H,13)\}\cup\{(J,13)\}\cup U \cup V$.
If $y$ assign the number $4$ white stone to a position outside the set $W$, the algorithm $T$ can help the black side to win the game by the following sequence:
\begin{itemize}
\item The black side captures $(I, 11)$ using stone $5$. After this, the black side will win if it capture $(J,11)$.
\item The white side captures $(J,11)$ using stone $6$. Otherwise, the black side can win the game by capturing $(J,11)$. 
\item The black side captures $(H, 10)$ using stone $7$. After this, the black side will win if it capture $(H,11)$. Because the white stones $2,4$ and $6$ cannot form an isosceles right triangle, the white side cannot win in the next step.
\item The white side captures $(H,11)$ using stone $8$. Otherwise, the black side can win the game by capturing $(H,11)$. 
\item The black side captures $(I, 9)$ using stone $9$. After this, the black side will win if it capture $(J,9)$ or $(H,9)$. Because none of three white stones can form an isosceles right triangle, the white side cannot win in the next step.
\item The white side captures a position using stone $10$. After this, at least one of $(J,9)$ or $(H,9)$ is empty.
\item The black side captures $(J,9)$ or $(H,9)$ using stone $11$ and wins the game, because either stone $1,3,9,11$ or $3,7,9,11$ is a square to win. One case of the sequence is shown in Figure \ref{f2}.
\end{itemize}
The main idea of the sequence is similar to checkmate in chess.
The black side tries its best to form a square without giving chance to the white side to do so.
To avoid failure, the white side has to stop the black side winning in the next step.
When the black side has two different positions to form a square in the next step, the white side can only capture at most one of them.
This leads to a win of the black side.
However, if $y$ assign the number $4$ white stone to a position inside the set $W$, the black side cannot be ensured a win using the above sequence.

\begin{algorithm}
\renewcommand{\algorithmicrequire}{\textbf{Input:}}
\renewcommand\algorithmicensure {\textbf{Output:} }
\caption{ $Z$}
\label{alg: Z}
\begin{algorithmic}[1]
\REQUIRE ~~
$W$;
black stone list $S_b$;
black stones $1,3$;
white stone list $S_w$;
board intersection recording steps $\Phi_{i,j}, i=A,...,S$, $j=1,...,19$;
number of maximal steps $m$;
\ENSURE ~~
Whether a win of the black side can always be ensured using this search $W_a$; number of cases explored $C_e$; number of cases where the black side can always be ensured a win using this search $C_w$; 

\STATE $C_e:=0$;
\STATE $C_w:=0$;
\FORALL{possible white stone $2$}
\FOR{$k=A,...,S$}
\FOR{$l=1,...,19$}
\FORALL{board intersection $\Phi_{i,j}, i=A,...,S$, $j=1,...,19$}
\STATE Assign $\Phi_{i,j}$ to empty;
\ENDFOR
\STATE Initialise $\Phi$ with stones $1,2,3$;
\STATE Initialise $S_b$ with stone $1$, then add stone $3$ from back;
\STATE Initialise $S_w$ with stone $2$;
\IF{ the intersection $\Phi_{k,l}$ has a stone}
\STATE Continue;
\ENDIF
\STATE $B_{case}:=false$;
\IF{ the intersection $\Phi_{k,l}$ is inside $W$}
\STATE $B_{case}:=true$;
\ENDIF
\IF{ $B_{case}$ is $false$}
\STATE Continue;
\ENDIF
\STATE Create the stone 4 at $(k,l)$;
\STATE Assign $\Phi_{k,l}$ to stone 4;
\STATE Add stone 4 to $S_w$ from back;
\STATE Create the array $S_n, n=5,...,m-1$;
\FORALL{$S_n, n=5,...,m-1$}
\STATE Assign $S_n$ to 0;
\ENDFOR
\FOR{$n=5,...,m-1$}
\IF{ $n$ is equal to $m-1$}
\STATE $n:=n-3$;
\STATE Remove the last stone from $S_b$ and assign the position of this stone to empty in $\Phi$;
\STATE Remove the last stone from $S_w$ and assign the position of this stone to empty in $\Phi$;
\STATE Continue;
\ENDIF

\algstore{myalg}
\end{algorithmic}
\end{algorithm}

\begin{algorithm}                     
\begin{algorithmic} [1]  
\algrestore{myalg}
\STATE $B_{next}:=false$;
\STATE $S_n:=S_n+1$;
\FORALL{$S_f, f=n+1,...,m-1$}
\STATE Assign $S_f$ to 0;
\ENDFOR
\STATE $P_{step}:=0$;
\IF{$n$ mod $2$ is $1$}
\IF{the black side can win in this step by adding a stone to form a square with 3 existing stones}
\STATE $C_w:=C_w+1$;
\STATE Print $n+1$;
\STATE Break;
\ENDIF
\FORALL{combinations of two unique black stones $p,q$ in $S_b$}
\IF{$p,q$ are on the same line of the board grid with a distance at most 3}
\FORALL{permutations of two empty positions $r,s$ that can form a square with $p,q$}
\STATE $P_{step}:=P_{step}+1$;
\IF{$P_{step}$ is equal to $S_n$}
\STATE Add a stone at $r$ to $S_b$ from back and update $\Phi$;
\STATE $B_{next}:=true$;
\STATE Break;
\ENDIF
\ENDFOR
\ENDIF
\IF{$B_{next}$ is $true$}
\STATE Break;
\ENDIF
\ENDFOR
\IF{$B_{next}$ is $true$}
\STATE Continue;
\ELSE
\STATE $n:=n-3$;
\STATE Remove the last stone from $S_b$ and assign the position of this stone to empty in $\Phi$;
\STATE Remove the last stone from $S_w$ and assign the position of this stone to empty in $\Phi$;
\STATE Continue;
\ENDIF
\ELSE
\IF{the white side can win in this step by adding a stone to form a square with 3 existing stones}
\STATE $n:=n-2$;
\algstore{myalg}
\end{algorithmic}
\end{algorithm}

\begin{algorithm}                     
\begin{algorithmic} [1]  
\algrestore{myalg}

\STATE Remove the last stone from $S_b$ and assign the position of this stone to empty in $\Phi$;
\STATE Continue;
\ENDIF
\IF{the black side can win in the next step by adding a stone to form a square with 3 existing stones}
\STATE Use a white stone to stop the black side winning, add this stone to $S_w$ from back and update $\Phi$;
\ENDIF
\ENDIF
\ENDFOR
\STATE  $C_e:=C_e+1$;
\ENDFOR
\ENDFOR
\ENDFOR
\IF{ $C_e$ is equal to $C_w$}
\STATE $W_a:=true$;
\ELSE
\STATE $W_a:=false$;
\ENDIF
\STATE Return $W_a, C_e, C_w$;
\end{algorithmic}
\end{algorithm}

\begin{figure}
\begin{center}
\includegraphics[width=0.9\textwidth]{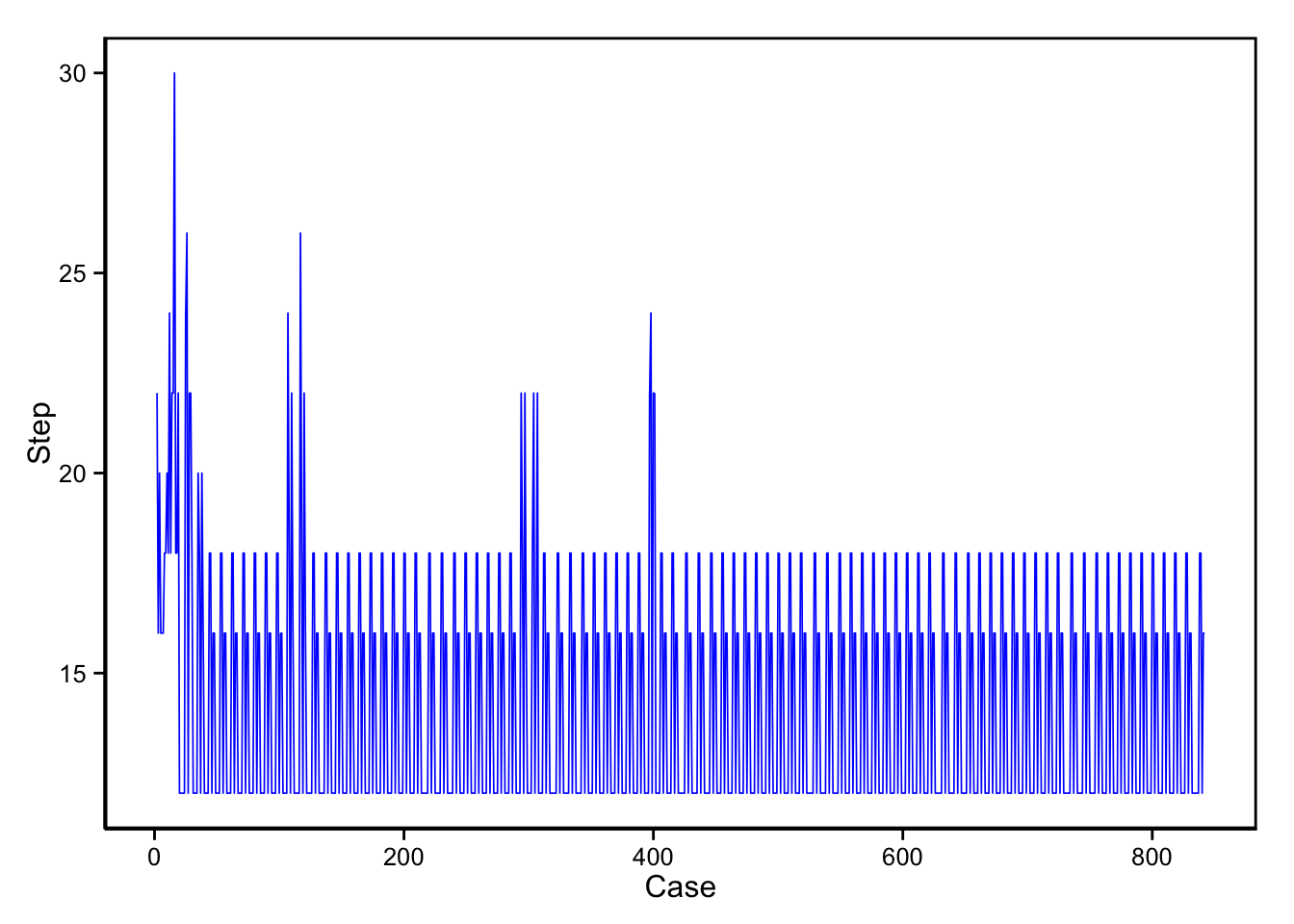}
  \caption{Results.}
  \label{f3}
  \end{center}
\end{figure}

Hence, we integrate the algorithm $T$ with the algorithm $Z$ to search for a sequence that ensures a win of the black side in any possible situation of the number $4$ white stone inside the set $W$. The algorithm $Z$ is a depth-first backtracking algorithm. It searches the tactics to ensure a win of the black side according to situations of the number $4$ white stone inside $W$. Since $W$ contains significantly fewer cases than the whole set $Y$, this algorithm is less complex than brute-force methods.

After the implementation using C++, we have obtained a positive answer from the algorithm $Z$.
It has ended with 842 cases in a short time, giving all positive outputs, as shown in Figure \ref{f3}.
Although the numbers of steps for the black side to win are not optimized, they are at most 30.

In summary, $\forall y \in Y$, we can always find $x \in X_0$ such that $F(x,y) = 1$ using the algorithm $T$.
This completes the proof.
\end{proof}

\bibliographystyle{plain}
 \bibliography{references}

\end{document}